\pdfoutput=1
\documentclass[11pt]{article}
\usepackage{acl}
\usepackage{times,latexsym}
\usepackage{url}
\usepackage[T1]{fontenc}
\usepackage{rotating}
\usepackage{multirow}
\usepackage{ulem}
\usepackage{comment}
\usepackage{subcaption}
\usepackage{url}
\usepackage{booktabs}
\usepackage{ulem}
\usepackage{amsmath}
\usepackage{latexsym}
\usepackage{multirow}
\usepackage{tabularx}
\usepackage{supertabular}
\usepackage{subcaption}
\usepackage{arydshln} 
\usepackage{booktabs} 
\usepackage{makecell} 
\usepackage{siunitx} 
\usepackage{booktabs,xcolor,siunitx}
\usepackage{colortbl} 
\definecolor{lightgray}{gray}{0.9}
\usepackage{xspace,mfirstuc,tabulary}

\title{Animate, or Inanimate, That is the Question for Large Language Models}

\author{\textbf{Leonardo Ranaldi, Giulia Pucci and Fabio Massimo Zanzotto} \\
 School of Informatics, University of Edinburgh, UK. \\
 	  Department of Computing Science, University of Aberdeen, UK \\Human-Centric ART Group, University of Rome Tor Vergata, Italy \\{
  {\tt [first\_name].[last\_name]@uniroma2.it},
  } }

\begin{document}
\maketitle
\begin{abstract}
The cognitive essence of humans is deeply intertwined with the concept of animacy, which plays an essential role in shaping their memory, vision, and multi-layered language understanding. Although animacy
appears in language via nuanced constraints on verbs and adjectives, it is also learned and refined through extralinguistic information. Similarly, we assume that the LLMs’ limited abilities to understand natural language when processing animacy are motivated by the fact that these models are trained exclusively on text.

Hence, the question this paper aims to answer arises: \textit{can LLMs, in their digital wisdom, process animacy in a similar way to what humans would do?} We then propose a systematic analysis via prompting approaches. In particular, we probe different LLMs by prompting them using animate, inanimate, usual, and stranger contexts. Results reveal that, although LLMs have been trained predominantly on textual data, they exhibit human-like behavior when faced with typical animate and inanimate entities in alignment with earlier studies. Hence, LLMs can adapt to understand unconventional situations by recognizing oddities as animated without needing to interface with unspoken cognitive triggers humans rely on to break down animations.
\end{abstract}

\section{Introduction}
The mnemonic abilities underlying cognitive processing seem to enable animate entities and concepts to be more easily memorized, which highlights the role of the animacy effect in human cognition \cite{doi:10.1073/pnas.0703913104,doi:10.1177/0956797613480803}.

Animacy is manifested via language through the faculty that humans have in using certain verbs or adjectives with animate and inanimate entities and accordingly inferring and reasoning about the mental states, intentions, and reactions of others. This allows them to navigate and understand social interactions. For this reason, using NLP models in increasingly complex social contexts necessitates the same ability to capture these socio-cognitive dynamics.

Current Large Language Models (LLMs) \cite{chowdhery2022palm,touvron2023llama,openai2023gpt4}, such as the GPTs \cite{openai2023gpt4}, PaLM \cite{chowdhery2022palm}, and Llama \cite{touvron2023llama},
are trained merely on textual data and cannot access non-verbal information, unlike humans. 
If faced with discerning animacy, they must infer it from its downstream linguistic implications, diverging from humans who also benefit from visual and physical stimuli. Thus, a fundamental question arises: \textit{Do LLMs perceive and respond to animacy hooks in language in a way as close to human comprehension as possible?}

\begin{figure}[t]
\centering
         \begin{minipage}{0.88\linewidth}
     \centering
     \includegraphics[width=\linewidth]{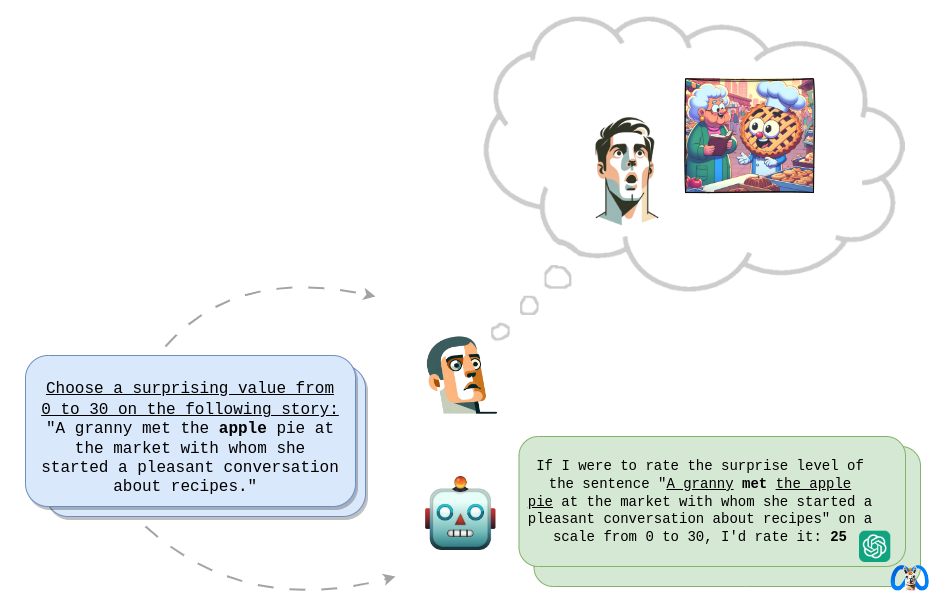}
   \end{minipage} 
   \caption{Large Language Models seem to be as surprised as humans at the thought of experiencing inanimate entities in animate contexts.} 
   \label{fig:our_work}

\end{figure}

This work investigates whether LLMs ``behave'' as humans when dealing with animacy.
We conduct extensive investigations using LLMs as subjects in psycholinguistic experiments developed for humans. Hence, we explore the behaviours of LLMs in answering to infractions of selective constraints associated with animacy in typical and atypical settings. 
Complementing the foundation work of \citet{warstadt-etal-2020-blimp-benchmark,spiliopoulou-etal-2022-events,hanna-etal-2023-language}, we study the animacy effect by operating via prompting approaches. We aim to elicit different LLMs to understand scenarios and situations that demand intricate reasoning passage. 
We discover that, like humans, LLMs generally prefer sentences adhering to animacy-related constraints, greatly preferring these constructions. These similarities are not strictly constrained to typical animacy; in fact, the behaviour of LLMs encountering atypical animate entities still seems to align with that of humans, both in terms of surprise at a first impression and downstream of adaptation by exhibiting significantly less surprise. 

Our findings can be summarized as follows:
\begin{itemize}
\item By proposing a systematic analysis based on the prompting of LLMs, we evaluate the animacy effect, extending the results obtained from previous contributions.
\item In particular, using psychological tests designed for humans, we observe that the LLMs not only prefer sentences that adhere to animacy constraints, as shown in \cite{warstadt-etal-2020-blimp-benchmark}, but are able to adapt awareness in atypical scenarios just as humans do \cite{10.1162/jocn.2006.18.7.1098}.
\item Finally, we demonstrate that LLMs not only demonstrate robust behaviour comparable with humans but also deliver answers that best approximate the placed expectations. 
\end{itemize}

\section{Models \& Methods}
\label{sec:models}

To investigate whether Large Language Models (LLMs) are able to understand and generate language in a way that reflects human expectations, we need to understand whether they are able to best approximate human knowledge of words and their cognitive passages. 
Hence, using LLMs as subjects~(\S~\ref{sec:subjects}), we study if behaviours can be manifested in human-designed experiments conceived for studying animacy~(\S~\ref{sec:human_exps}).
We propose a systematic prompt-based approach for LLMs through which we discuss the results in \S~\ref{sec:human_exps_on_LLMsubjects}. Finally, we outline a general discussion of the findings in \S~\ref{sec:gen_disc}.

\subsection{The Subjects}
\label{sec:subjects}

The animacy effects behind state-of-the-art Large Language Models are analyzed via systematic prompting in three groups of models: 
\begin{itemize}
\itemsep0em 
    \item two subjects from the OpenAI family \cite{openai2023gpt4}: GPT-3.5 and GPT-4;
    \item three subjects form the Meta family \cite{touvron2023llama}: Llama2-chat-7b, -13b, -70b;
    \item two subjects form the Mistral family \cite{jiang2023mistral,jiang2024mixtral}: Mixtral8x7b, Mistral-7b ;
\end{itemize}

We use both open-source models\footnote{To simplify the discussion, we omit "chat", "b". The resulting names are Llama2-7, -13, -70  Mixtral, and Mistral-7} to make our work more reproducible and closed-source models because they demonstrate outstanding performance in many NLP tasks. 

Finally, as we describe in each experiment, we evaluate the accuracy scores computed via string matching between the final and the target answers (detailed information in Appendices \ref{app:model_versions}, \ref{app:model_info}).

\subsection{Selected Experimental Settings}
\label{sec:human_exps}

To adapt our `subjects' to the experimental settings proposed on humans, we discern between two different types of experiments: (1) typical animacy (\S~\ref{sec:Typical_Animacy}); (2) atypical animacy (\S~\ref{sec:atypical_animacy}). The two different kinds of experiments are needed as typical animacy is more a lexical task, and atypical animacy is more a contextual task from the point of view of LLMs.

\subsubsection{Typical Animacy}
\label{sec:Typical_Animacy}

In typical animacy experiments, subjects are prompted to determine which word in a pair is animate and which is not (e.g., if "frogs" are animated and "mountains" are not). Hence, we use two different settings: 
\begin{itemize}
\itemsep0em 
    \item the Benchmark of Linguistic Minimal Pairs (BLiMP) \cite{warstadt-etal-2020-blimp-benchmark} in \S~\ref{sec:Typical_Animacy_exp};
    \item the Benchmark of Sentence Plausibility (BSP) \cite{VEGAMENDOZA2021107724} in \S~\ref{sec:Typical_Animacy_exp1.2};
\end{itemize}

\begin{table}[h]
\small
\centering 
 \begin{tabular}{c|c}
 \textbf{Acceptable}  &  \textbf{Example}  \\
\hline
\multicolumn{2}{c}{\textbf{Sub-task}: Passive}  \\
\hline
\texttt{Yes}  & The glove was noticed by some \textit{woman}.	  \\
\texttt{No}  & The glove was noticed by some \textit{mouse}. \\
\texttt{Yes}  & Galileo is concealed by the \textit{woman}.	  \\
\texttt{No}  & Galileo is concealed by the \textit{horse}.  \\
\hline
\multicolumn{2}{c}{\textbf{Sub-task}: Transitive}  \\
\hline
\texttt{Yes}  & \textit{Beth} scares Roger.  \\
\texttt{No}  & \textit{A carriage} scares Roger.  \\
\texttt{Yes}  & \textit{Tanya} admires Melanie.  \\
\texttt{No}  & \textit{Music} admires Melanie.  \\
\end{tabular}
\caption{Two examples from the \textit{Transitive} and \textit{Passive} datasets. Each is a minimal pair of sentences: one Acceptable \texttt{(Yes)} and one not \texttt{(No)}.}
\label{tab:BLiMP_examples}
\end{table}

In BLiMP, we select two sub-tasks: \textit{transitive-animate} and \textit{passive-animate}. Each sub-task has 1,000 pairs of synthetic English sentences that are similar but differ by only one/two words~(Table~\ref{tab:BLiMP_examples}). 

Meanwhile, in BSP, we use sentences containing plausible and implausible words with different nuances. The resource contains 1,500 synthetic sentences in English. Each sentence has a fixed initial part and an interchangeable final part between animated plausible, animated implausible (inherent and non-related), and inanimate implausible (inherent and non-related) words~(Table~\ref{tab:BSP_examples}). From the point of view of our subjects, that is, LLMs, this psychological experiment is translated into a lexical task. 

\begin{table}[h]
\small
\centering 
 \begin{tabular}{l r}
 \hline
 \multicolumn{2}{l}{Sentence:} \\ 
\multicolumn{2}{c}{\textbf{At the club the cocktails are served by the \_}} \\ 
\hline
\multicolumn{2}{c}{\textbf{\textit{Plausible}}}	
\\
\hdashline
Control  & barmaid	  \\
\multicolumn{2}{c}{\textbf{\textit{Implausible}}}	  \\
\hdashline
Animate-Related  & drunkard  \\
Animate-Unrelated  & queen	  \\
Inanimate-Related  & tonic  \\
Inanimate-Unrelated  & dirt  \\
\hline
\end{tabular}
\caption{Example from Benchmark of Sentence Plausibility. Each sentence has a \textit{plausible} and four \textit{non-plausible} words. As proposed by \citet{VEGAMENDOZA2021107724} we use the options as different tasks.}
\label{tab:BSP_examples}
\end{table}

\subsubsection{Atypical Animacy}
\label{sec:atypical_animacy}

In contrast to \S~\ref{sec:Typical_Animacy}, for investigating if the subjects are able to detect animacy without relying on lexical information of the target word, we employ repetition and contextual study where the inanimate entities are treated as animated entities \cite{10.1162/jocn.2006.18.7.1098}. This shifts the focus from the lexical knowledge of the target word to the contextual knowledge. The human experiments are based on N400, a brain response measured by EEG that rises when processing semantically anomalous input. 

The repetition study measured participants' N400 responses while reading cartoon-like stories in which a typically inanimate entity behaved as animate (Table \ref{tab:N400_examples}).
\citet{10.1162/jocn.2006.18.7.1098} found that although initially surprised by the atypically animated entity, participants quickly adapted, producing increasingly lower N400 responses. 
In contrast, the contextual study performs the measures only behind a contextualization part since the repetition experiment shows similarities with the work of \citet{10.1162/089892998563752}. These contexts are given as in Table \ref{tab:N400_context_examples} where people are asked to read the story with one of the targets alternatively.

\begin{table}[h]
\small
\centering 
 \begin{tabular}{l}
\hline
A girl told a sandwich that an attack was imminent. \\
The sandwich wailed that his family was in danger. \\
The girl told the sandwich that public places were \\
the most dangerous. The sandwich immediately\\
started calling everyone he knew. The sandwich \\
was [\ \textbf{targets} ]\ and 
wanted to make sure none \\
of his loved ones were in danger \\
\hline
targets: \textbf{delicious}, \textbf{worried}  \\
\hline
\end{tabular}

\caption{Example from translated \textit{context story} of N400 \cite{10.1162/jocn.2006.18.7.1098}.}
\label{tab:N400_context_examples}
\end{table}

Hence, these experiments are useful for investigating the following questions: \textit{Can LLMs adapt to animated entities at the token level despite being typically inanimate? Or is animate processing limited to a simple type-level understanding?} We replicate these with LLMs to answer this question, using their surprise to model N400 responses.

We conduct two different experiments:
in the first experiment presented in \S~\ref{sec:atypical_animacy_exp1}, we reproduce the repetition and context as in \cite{10.1162/jocn.2006.18.7.1098}; in a second experiment (\S~\ref{sec:atypical_animacy_exp2}), we analyze the impact of context adaptation as proposed in \cite{Boudewyn2019}. For clarity, we introduce the original study and the methods we used for the context adaptation of LLMs. Finally, we report our empirical results and compare them with those of the original study.

\begin{table*}[h]
\centering 
 \begin{tabular}{c}
 \hline

\hline
($T_1$) A granny met the (\textbf{confectioner}-\textbf{apple pie}) at the market with whom she started a pleasant conversation \\
about recipes. 
($T_2$) The (\textbf{confectioner}-\textbf{apple pie}) confided a 
secret recipe to the granny. ($T_3$) But the granny  \\
deceived the (\textbf{confectioner}-\textbf{apple pie}) by making off with the recipe herself. 
($T_4$) The (\textbf{confectioner}-\textbf{apple pie}) \\
discovered the deception and wanted to reprimand the granny. But the granny pleased the  \\
(\textbf{confectioner}-\textbf{apple pie}) with an (\textbf{confectioner}-\textbf{apple pie}) with an even better version of the recipe. \\
($T_5$) The (\textbf{confectioner}-\textbf{apple pie}) understood that this was the ultimate recipe and  
apologized for the \\
misplaced distrust.  \\
\hline

\end{tabular}

\caption{Example from translated version of N400 \cite{10.1162/jocn.2006.18.7.1098}. The first tokens indicate an \textit{acceptable} example, and the numbers indicate the sentences given as context.}
\label{tab:N400_examples}
\end{table*}

\subsection{Experimenting with LLM subjects}
\label{sec:human_exps_on_LLMsubjects}

\subsubsection{Experiment 1: Typical Animacy on BLiMP}
\label{sec:Typical_Animacy_exp}

\paragraph{Prompt definition}
By constructing a series of prompts over the datasets presented in \S \ref{sec:Typical_Animacy}, we test the models' answers to animacy in situations where the animacy of an instance aligns with its more general type.

Sentence pairs in BLiMP \cite{warstadt-etal-2020-blimp-benchmark} is built as follows: one sentence respects the animacy constraints, and the other violates them. Hence, there is a straightforward way to evaluate the LLM's ability to surpass the animacy test. We ask them to answer the following prompt: 
\begin{center}
\small
{\setlength{\fboxsep}{-1.2pt} 
\colorbox{lightgray}{ 
    \begin{tabular}{p{0.94\linewidth}}
          \texttt{\textbf{Choose which example is acceptable between A and B.}} \\
\texttt{A) Galileo is concealed by the \textit{woman}.} \\
\texttt{B) Galileo is concealed by the \textit{horse}.} \\
\texttt{\textbf{Answer:}} \\
          
    \end{tabular} 
        } 
    }
\hfill
\end{center}

A model gets a correct example if it chooses a sentence that respects the animacy constraint. 

Following this approach, we evaluate the accuracies by performing a string matching- between the generated answers and the target values on both sub-tasks.

\begin{figure}[h]
\centering
         \begin{minipage}{0.8\linewidth}
     \centering
     \includegraphics[width=\linewidth]{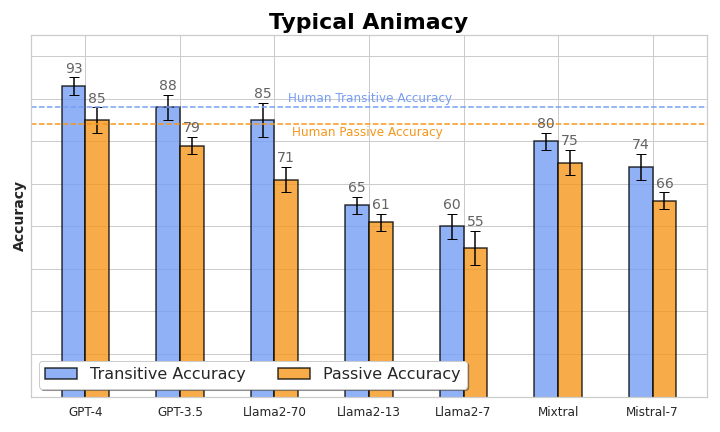}
   \end{minipage}
   \caption{Large Language Models performances on animate- transitive and passive sub-tasks of BLiMP benchmark \cite{warstadt-etal-2020-blimp-benchmark}.} 
   \label{fig:blimp}

\end{figure}

\paragraph{Results}
The results of this first experiment set are intriguing. The OpenAI family behave on par with respect to humans, and the Meta and Mistral families are catching up. Figure \ref{fig:blimp} shows the results of each model (vertical bars) and the results obtained by humans (horizontal dashed lines), as presented in \cite{warstadt-etal-2020-blimp-benchmark}. The accuracy metric has to be intended as the percentage of examples in which human or artificial subjects preferred the acceptable sentence of the given pair.

Human Transitive Accuracy is reached by GPT-3.5 and topped by GPT-4. This seems to suggest that these models can handle the lexicon to determine typical animacy. Llama lags behind, but it is reaching the Human level. 

Similarly, in the passive scenario, GPT-4 performs very close to humans. 
Concerning GPT-3.5, Llama2-70 and Mixtral have comparable and slightly lower performance in the transitive scenario and significantly lower performance in the passive scenario, respectively.
Finally, the smaller models, i.e., with fewer parameters, underperform humans with average gaps of 20 points.

This difference between the transitive and passive case may be due more to differences in setting than to different animacy processing in the two scenarios. 
However, emerges that the composition of the choices is strongly class-related. Indeed, in the passive case, the target word, i.e., the most influential one, is always in the last position. 
In contrast, the target word is not the last token in the transitory case. Thus, heuristics related to the sensitivity of the choices in the input prompts' structure may be present.

\begin{figure*}[h]
\centering
         \begin{minipage}{0.38\linewidth}
     \centering
     \includegraphics[width=\linewidth]{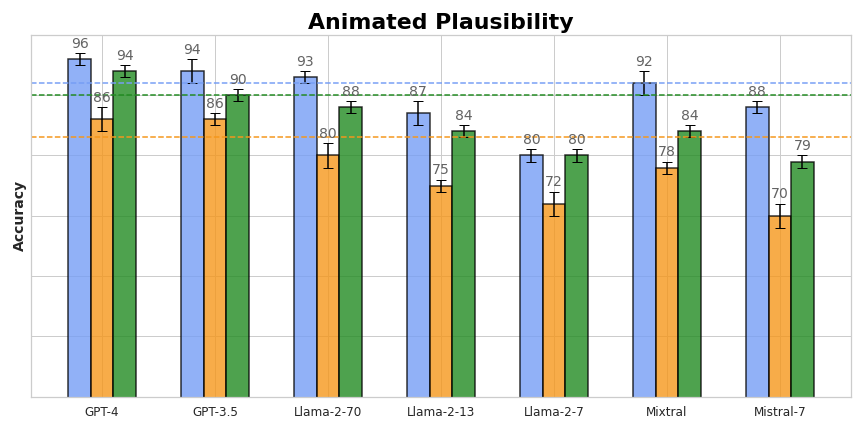}
   \end{minipage}
            \begin{minipage}{0.38\linewidth}
     \centering
     \includegraphics[width=\linewidth]{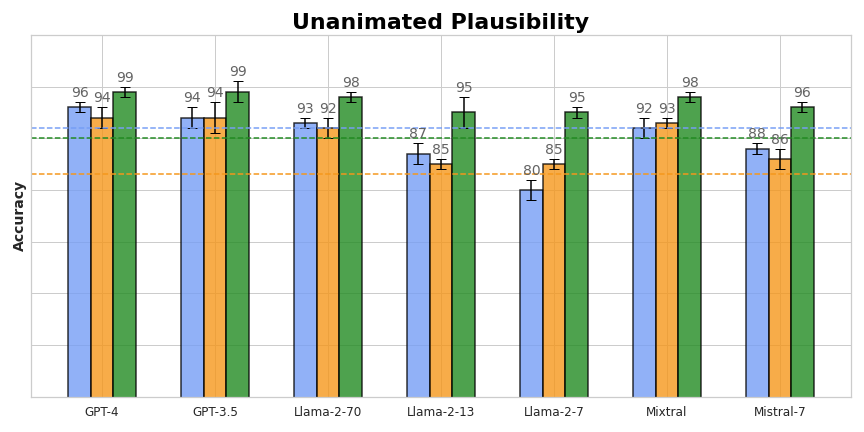}
   \end{minipage}
   \begin{minipage}{0.4\linewidth}
     \centering

     \includegraphics[width=\linewidth]{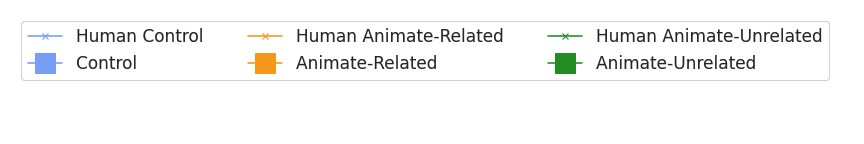}
   \end{minipage}
      \begin{minipage}{0.4\linewidth}
     \centering
     \includegraphics[width=\linewidth]{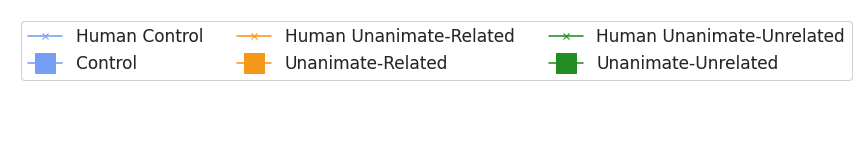}
   \end{minipage}
   \hfill   
   \caption{Large Language Models performances on the BSP benchmark \cite{VEGAMENDOZA2021107724}.} 
   \label{fig:BSP}

\end{figure*}

\subsubsection{Experiment 2: Typical Animacy on BSP}
\label{sec:Typical_Animacy_exp1.2}
\paragraph{Prompt definition}
In the second Experiment, we structured the prompting phase similarly. Hence, by constructing a series of prompts over the BSP benchmark (\S~\ref{sec:Typical_Animacy}), we test the model's responses in situations where plausible and implausible sentences with animated and non-animated components were provided. 

Following \citet{VEGAMENDOZA2021107724}, we analyze the model's answers to the plausibility question on five inputs constructed from a sentence and completed with different types of answers.
Hence, there is a straightforward way to evaluate our subject's ability to surpass the plausibility test. 
Hence, given the following prompt: 

\begin{center}
\small
{\setlength{\fboxsep}{-1.2pt} 
\colorbox{lightgray}{ 
    \begin{tabular}{p{0.95\linewidth}}
          \texttt{\textbf{Is the following sentence plausible? Answer by choosing (Yes) or (No). }} \\
\texttt{\textbf{Sentence:} In ancient Egypt the people were governed by the \uline{pyramid}.} \\
\texttt{\textbf{Answer:}} \\
          
    \end{tabular}
        } 
    }
\hfill
\end{center}

A model gets an answer if it answers the question with \textit{Yes}/ \textit{No}, respecting plausibility. 
The accuracy of the LLMs is computed in this way: for the plausible control word, the accuracy counts the percentage of \textit{Yes}, and for all the implausible words the percentage of \textit{No}.     

\paragraph{Results}
In this second experiment, LLMs of the OpenAI family behave similarly to humans as in the previous. Figure \ref{fig:BSP} shows the accuracy results of each model and the results obtained by humans, as presented in \cite{VEGAMENDOZA2021107724}. 
Dealing with animated words (see Figure \ref{fig:BSP}), humans and LLMs behave similarly. Indeed, humans perform on animate-unrelated similarly to how they perform on control words. Instead, they are less able to recognize animate-related as making target sentences implausible. The same happens for all the LLM subjects. GPT-4 performs better than humans, and it keeps the difference in recognizing the implausibility of sentences built with animated-unrelated and animated-related words.
Moreover, when dealing with unanimated words (see the right plot in Figure \ref{fig:BSP}), humans and LLMs behave similarly. Humans recognize the implausibility of unanimated-unrelated words but have a slight decrease in recognizing unanimated-related ones. The same trend happens for all the LLM subjects and, consistently in other experiments, the OpenAI family performs better than humans.

In humans, these differences in the plausibility of animate and non-animate cases are given by a combination of cognitive factors, as explained by \citet{VEGAMENDOZA2021107724}.
Consequently, as in the experiments in \S~\ref{sec:Typical_Animacy_exp}, GPTs perform comparably to humans and sometimes outperform.
However, even in this task, there is a robust structural component related to fearfulness. The target words, i.e., those that provide the final decision, are always in the last position. 
Therefore, there may be a heuristic related to the sensitivity of the structure of choices in the input prompts.

\subsubsection{Experiment 3: Atypical Animacy - Repetition}
\label{sec:atypical_animacy_exp1}

\paragraph{Human experiment and its results} The repetition experiment on Atypical Animacy \cite{10.1162/jocn.2006.18.7.1098} measure the N400 responses of a series of participants who listened to Dutch stories containing a typical animate or an inanimate entities behaving as if it were a human being. The N400 values are measured in three stages: the first ($T_1$), the third ($T_3$), and the fifth ($T_5$) mention of the entity (see, for example, Table \ref{tab:N400_examples} with confectioner and apple pie). \citet{10.1162/jocn.2006.18.7.1098} discovered that: 
\begin{itemize}
\itemsep0em 
    \item in the case of animated entities, participants have a moderate N400 response to the first mention ($T_1$) and a low response to subsequent mentions ($T_2$ and $T_3$);
    \item in the case of inanimated entities, participants initially ($T_1$) have a high N400 response to the atypically animated entity, and, as the mentions progress ($T_2$ and $T_3$), their N400 responses are very close to the responses from the mentions of the animated entity. 
\end{itemize}
Thus, while the humans are initially surprised by the atypically animated entity, they quickly adapt to the situation and no longer find it surprising. Moreover, they show that responses do not derive from lexical repetition but from context. In fact, in the contextual experiment, they provide a context. Only at the end did they estimate N400 responses of the participants obtaining low results for inanimate entities in atypical inanimate contexts.

\paragraph{Prompt definition}
To estimate a surprise value analogous to the N400, state-of-the-art studies examine token probability values. However, some of the models used in our study do not provide access to probability values, prompting us to define a series of prompts to query the model about its level of surprise systematically. In particular, for each of the 60 examples, we estimate the surprise of the animate and inanimate entity given by the context at each time-step, denoted as $T_n$, (which, in our case, refers to an input-prompt). For instance, to model the inanimate N400 response at $T_1$ in the example from Table \ref{tab:N400_examples}, we construct the following prompt: 

\begin{center}
\small
{\setlength{\fboxsep}{-1.2pt} 
\colorbox{lightgray}{ 
    \begin{tabular}{p{0.95\linewidth}}
          \texttt{\textbf{Choose a surprising value from 0 to 30 on the following story:}} \\
\texttt{A granny met the \uline{apple pie} at the market with whom she started a pleasant conversation about recipes.} \\
\texttt{\textbf{Answer:}[num]} \\
          
    \end{tabular}
        } 
    }
\hfill
\end{center}

Following the time-steps, we introduce additional prompts by contextualizing the preceding story, whether animate or inanimate. For example, for the inanimate scenario:

\begin{center}
\small
{\setlength{\fboxsep}{-1.2pt} 
\colorbox{lightgray}{ 
        \begin{tabular}{p{0.95\linewidth}}
    \texttt{\textbf{Given the following story:}} \\
\texttt{A granny met the \uline{apple pie} at the} \\
\texttt{market with whom she started} \\
\texttt{a pleasant conversation about recipes.} \\
\texttt{The \uline{apple pie} confided to.............} \\
\texttt{\textbf{Choose a surprising value from 0 to 30 on the following story:}} \\
\texttt{The \uline{apple pie} that this was the ultimate recipe and apologized for the misplaced} \\
\texttt{distrust.} \\
\texttt{\textbf{Answer:}[num]} \\
          
    \end{tabular}
        } 
    }
\hfill
\end{center}

Hence, we compute the average surprise value of examples containing animate and inanimate entities separately at each time-step.

\begin{figure}[h]
\centering
         \begin{minipage}{0.7\linewidth}
     \centering
\includegraphics[width=\linewidth]{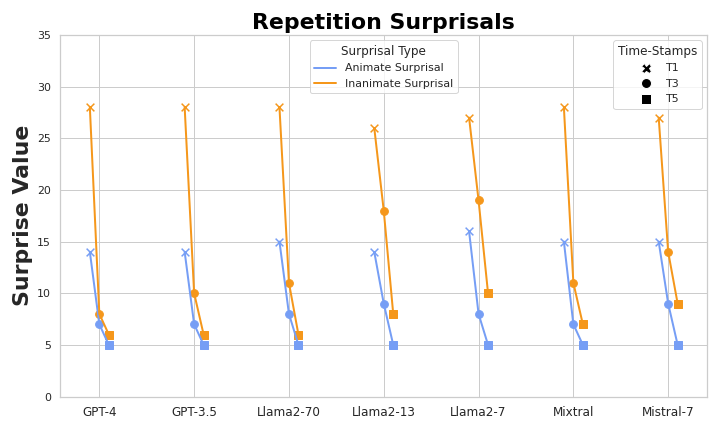}
   \end{minipage}
   \caption{Average surprise values provided by the LLMs at input-prompt $T_1$, $T_3$ and finally $T_5$.} 
   \label{fig:exps2}

\end{figure}

\paragraph{Results}
The LLMs follow the general trends of human N400 responses (Figure \ref{fig:exps2}). Indeed, as reported in \S~\ref{sec:atypical_animacy_exp1}, human N400 responses for animate and inanimate critical words diverge at $T_1$ and come closer at $T_3$ and $T_5$. LLM subjects behave similarly. In fact, at $T_1$, models are surprised by the inanimate entity and trimmed by the animate one. At later steps ($T_3$ and $T_5)$, the surprises of inanimate entities decrease until they reach levels similar to animate entities. LLMs seem to adapt, just as humans do. However, the raw results do not show that the models adapt to the same extent as humans. 

We use the Wilcoxon signed-rank test for making the experiments robust as \cite{10.1162/jocn.2006.18.7.1098}. We observe distinct surprise values at each time step. As with humans, LLMs have a statistically significant difference between the surprises of animate and inanimate entities, for example, $T_1$ of Figure \ref{fig:p_value}.
However, while there is no difference between humans at $T_3$, there are differences (p < 0.01) in most models; only the largest do not have any. At $T_5$, differences disappear only in the large models. Although the models can generally approximate human N400 responses to atypical animacy trends, only the most significant and most potent fully replicate human adaptation.

\begin{table}[h]
\begin{center}
         \begin{minipage}{0.75\linewidth}
     \centering
\includegraphics[width=\linewidth]{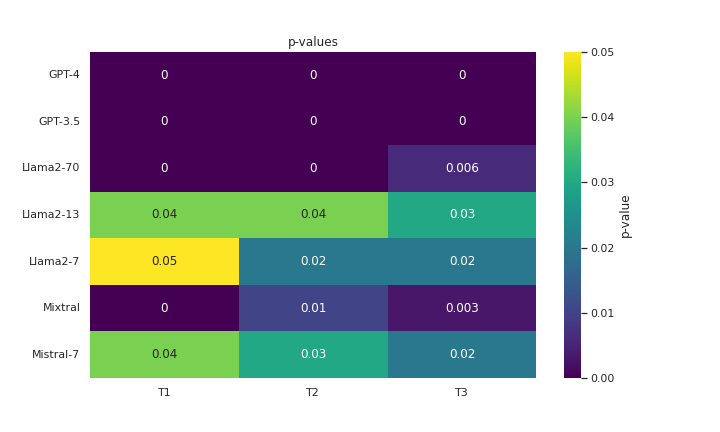}
   \end{minipage}
   \hfill   
   \caption{Statistical significance of the difference between animate and inanimate surprisal, by LLMs response values and time-step} 
   \label{fig:p_value}
\end{center}
\end{table}

\subsubsection{Experiment 4: Atypical Animacy - Context Experiment}
\label{sec:atypical_animacy_exp2}

\paragraph{Human experiments \& results}
\citet{10.1162/jocn.2006.18.7.1098} discover that contextual appropriateness seems to neutralize animacy violations, that is, non-appropriate adjectives (such as \textit{``worried''}, for example, in Table \ref{tab:N400_context_examples}) are not generating much surprise if the context suggests them. Moreover, context can even make an animacy-violating predicate more preferred than an animacy-obeying canonical predicate if the context justifies this. 
\paragraph{Prompt definition}
By using examples for the context experiment (as in Table \ref{tab:N400_context_examples}), we ask for a surprising value for animate and inanimate adjectives for each of the 60 stories proposed by \citet{10.1162/jocn.2006.18.7.1098}. 
In particular, we use input-prompt structures closer to the previous:

\begin{center}
\small
{\setlength{\fboxsep}{-1.2pt} 
\colorbox{lightgray}{ 
    \begin{tabular}{llp{6.5cm}} 
&&\texttt{\textbf{Given the following context:}} \\
&\multirow{1}{*}{\mbox{\begin{turn}{90}+Context\end{turn}}}&\texttt{A girl told a sandwich that an attack was imminent. The sandwich wailed that his family was in danger. The girl told the sandwich that public places were the most dangerous. The sandwich immediately started calling everyone he knew.}\\
\hline
&\multirow{2}{*}{\begin{turn}{90}Baseline\end{turn}}&\texttt{\textbf{Choose a surprising value from 0 to 30 on the following story:}} \\
&&\texttt{The sandwich was \uline{delicious} and wanted to make sure none of his loved ones were in danger.}\\
&&\texttt{\textbf{Answer:}[num]} \\
    \end{tabular}
            } 
    }
\hfill
\end{center}

To estimate absolute values, we also ask for baseline surprises, that is, those of the inanimate adjective without the context of the whole story.

\begin{figure}[h]
\centering
         \begin{minipage}{0.8\linewidth}
     \centering
\includegraphics[width=\linewidth]{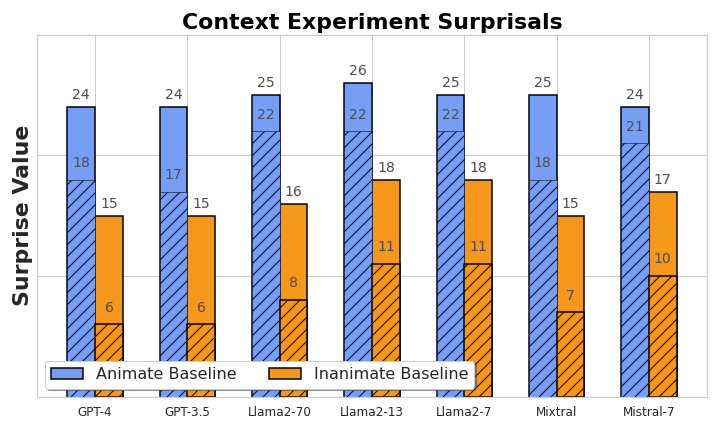}
   \end{minipage}
   \caption{Average surprise values.} 
   \label{fig:exp3}

\end{figure}

\paragraph{Results}
Even in this experiment in Figure \ref{fig:exp3}, LLM subjects behave similarly to human subjects. The animate baseline is larger for all subjects than the inanimate baseline. Even in the baseline, there is conflicting information as to the presence or absence of animacy of the selected subject. Moreover, the surprise drop with the context is more significant with inanimated than animated adjectives. This is in line with the human experiments.

\subsection{General discussion}
\label{sec:gen_disc}

The experiments deliver a coherent message: \textit{despite their lack of embodiment and senses, LLMs behave as humans in animacy understanding. }
For the two typical animacy tasks (Experiments 1 and 2), LLMs can be surprised by both animate and inanimate examples like humans. The intriguing part is that LLMs of the OpenAI family behave very similarly to humans. Moreover, the atypical animacy tasks (Experiments 3 and 4) clear the ground from the fact that LLMs possibly use only lexical information correlated to words. Indeed, in the atypical animacy tasks, subjects are investigated on their level of surprise over an inanimated object performing actions as if they were animated. These two experiments aim to track how surprise lowers when subjects are exposed to more context. Hence, this is not a lexical task. Nevertheless, similar to the results of typical animacy, LLM subjects behave as human subjects: their level of surprise lowers more for inanimated objects than animated ones.   
Finally, we complete our analysis by providing evidence of the stability of the assessments on the generations in Appendix~\ref{app:error_analysis}.

\section{Related Work}
The relationship between animacy and language has long been a subject of interest in cognitive linguistics. Visualized as a spectrum, animacy influences linguistic structures and shapes cognitive interpretations of entities based on their perceived liveliness. There is a wide path to studying animacy in Natural Language (\S~\ref{sec:animacy_natural_language}). At the same time, these explorations began to spread within the NLP community with attractive evidence of the abilities of Language Models in handling linguistic properties (\S~\ref{sec:LLMs_back}).
However, the advent of Large Language Models (LLMs) has revolutionized many previously adopted analytical methods. Therefore, we ask whether ongoing models can handle linguistic properties in the same way as their predecessors. Even more so whether these models are able to generate answers in line with the expectations placed on them by humans. In an analytical scenario, we adopt a psycholinguistic lens, treating these models as subjects to assess their understanding and processing of animacy.

\subsection{The Animacy in Natural Language}
\label{sec:animacy_natural_language}
Animacy in cognitive processes is described as going along a continuum \cite{GarcaGarca2018ShiftingFA}. This is shown through a hierarchy of humans, animals, and objects in language. Entities are distinguished based on their position in this ranking through sentence structure or word form \cite{Gass1984ARO}.
It can be traced at the broad category and the specific instance levels. Similarly, linguistic animacy is not solely grounded in biological factors but also hinges on the speaker's emotional connection and empathy towards a specific entity \cite{VihmanNelson+2019+260+267}. 
The impact of animacy in language is not uniform across different languages; it can range from explicit markers of animacy to more subtle influences.
Such subtleties encompass strict constraints based on animacy \cite{Caplan1994,Buckle2017} and nuanced grammatical impacts \cite{Rosenbach2008}. For instance, sentences more frequently begin with animate entities, even if this results in less conventional structures \cite{FERREIRA1994715,Fairclough2008}. 
We analyzed the distinction between humans and inanimate objects and the constraints based on animacy. Such a pronounced differentiation is anticipated to yield more discernible effects in LLMs.

\subsection{Large Language Models as Tests Subjects}
\label{sec:LLMs_back}

Previous works present investigations on Language Models\footnote{non-large and non-instruction-tuned or further refined} linguistic capabilities on structural properties and operating through the analysis of assigned probabilities of given sentences. Such techniques have previously been used to deepen LMs' understanding of constructs such as negation, structural agreement, and in-context priming \cite{sinclair-etal-2022-structural,jumelet2024languagemodelsexhibithumanlike}.
Recent studies have compared the performance of LLMs with human cognition by utilizing surprisal, the negative log probability of a sequence, as an estimator for cognitive exertion \cite{michaelov2023peanuts,hanna-etal-2023-language}. LLM has displayed notable versatility, demonstrating significant correlations with human values \cite{Aurnhammer2018ComparingGA,goodkind-bicknell-2018-predictive,truong-etal-2023-language}. 
\paragraph{Large Language Models}
Compared with the smaller Language Models, ongoing Large LMs (GPTs \cite{openai2023gpt4}, Llamas \cite{touvron2023llama}, \textit{et alia}) have been demonstrating capabilities in challenging complex tasks by delivering multi-step reasoned answers \cite{wei2022emergent,wei2023chainofthought}.
The refinement techniques employed to lead the models to best approximate human expectations in their responses are increasingly operated \textit{(e.g., instruction-tuning \cite{ouyang2022training}, reinforcement learning from human preferences \cite{christiano2023deepreinforcementlearninghuman})}. 
We analyzed whether these methods that give rise to emergent capabilities also arise in psycholinguistic tasks applied to humans in detecting animacy. By complementing the foundation works \cite{michaelov2023peanuts,truong-etal-2023-language,spiliopoulou-etal-2022-events,buijtelaar-pezzelle-2023-psycholinguistic,hanna-etal-2023-language}, we extended the analyses to include further tests and revisited prompt-based evaluation by eliciting the models to generate responses (\S~\ref{sec:our_contribution}).
We evaluate different LLMs by employing them as subjects within a psycholinguistic framework, an increasingly adopted methodology in the field. This approach psychoanalyses the LLMs by evaluating their delivered responses to several questions, just as would be done to humans.

\subsection{Animacy in Large Language Models}

Previous works have investigated the LMs' abilities in processing animacy.
\citet{warstadt-etal-2020-blimp-benchmark} explored the phenomenon within the BLiMP framework. \citet{kauf2023event} investigate within LMs' overall event knowledge, concluding that models are adept at discerning animacy concerning selective constraints.
Our study goes forward exploring different animacy. We introduce a surprise score following \citet{10.1162/jocn.2006.18.7.1098} and \citet{michaelov2023peanuts,hanna-etal-2023-language}.
Then, we focused on transferring the experimental setting into generative models where accessing the internal weights (closed-source LLMs) is often impossible. These efforts aim to study scenarios in which the models efficiently capture the trends manifested by analyzing open LMs. Furthermore, we aim to investigate the correlation between the robustness and predictive accuracy of an LM by evaluating a broad spectrum of LLMs.

\subsection{Our Contribution}
\label{sec:our_contribution}

Completing the earlier foundational work (see \S \ref{sec:LLMs_back}) comparatively, our work goes beyond by:\\
\textbf{(\textit{i})} We propose a systematic prompting pattern and analyzing natural language responses as humans would. Specifically, we establish a promting pipeline for estimating the LLMs' understanding of the acceptability and plausibility of concepts related to animate and inanimate entities. Moreover, we extend to generative-based models an approach based on a series of progressive in-context prompts to simulate the estimation of the N400 neurological response.
\textbf{(\textit{ii})} Hence, by placing LLMs in atypical contexts with animated entities, we have shown similarities to the results of tests performed on humans and the results from previous contributions. \textbf{(\textit{iii})} Finally, we show that the prompting approaches are affected by minor bias that allows fair analogies between results obtained by LLMs and prior findings (Appendix~\ref{app:error_analysis}).

\section{Conclusion}
Large Language Models (LLMs) reveal capacities to solve repetitive cognitive tasks by exploiting them better and faster than humans.
We treat LLMs as subjects in psychological experiments, exploring whether they behave as humans when dealing with the concept of animacy. 
We show that LLMs behave as humans even when lexical information does not entirely lead the decision. This is astonishing, as humans' cognitive underpinnings are intricately linked with the concept of animacy. 
Although the LLMs subjects of our analysis are trained primarily on textual data without the support of extralinguistic information,
they show human-like behaviour when exposed to typical entities and adapted to unconventional narratives. The ability to adapt, while remarkable, is sometimes aligned with human fluidity. 
Hence, it becomes imperative to provide robust analyses of LLMs' behaviour in intricate social scenarios by studying the ability to emulate human-like processes. 
While this analysis shows significant results, current models are still grappling with assessing the understanding of social interactions. To truly build models aligned with human minds, they must merge their vast textual knowledge with a deep understanding of human social dynamics.

\section*{Limitations \& Future Works}
In this study, we used a series of behavioral experiments. Due to the ease of interaction with Large Language Models on benchmarks, they are suitable for comparing the models with human data. Although they have shown exciting features, analyzing the causal mechanisms by which these models processed the animated sentences is impossible. To analyze them, it would be appropriate to look at the integer weights of the models, which are not always open-source, as in the GPTs thing \cite{openai2023gpt4}. 

In future developments, we plan to extend the analysis to more languages to assess whether the models respond in the same way in similar scenarios and contexts involving languages beyond English. In addition, it is of interest to us to assess the impact of the in-context prompt, in particular, the degree to which the composition of the prompt may influence causal generations as done for the sycophantic behavioural study in \cite{ranaldi2024largelanguagemodelscontradict}. Last but not least, it will be of interest for us to analyze the internal dynamics that support the models' decisions in order to better understand the neuronal patterns that motivate the generations \cite{mohebbi-etal-2024-transformer}.

\section*{Ethics Statemet}
In our work, ethical topics were not addressed. The data comes from open-source benchmarks, and statistics on language differences in commonly used pre-training data were obtained from official sources without touching on gender, sex, or race differences.

\bibliography{p2021,custom}

\appendix

\clearpage

\begin{table}[t]
\section{Error Analysis}
\label{app:error_analysis}
Although we have observed human-like behaviors as extensively discussed in the previous sections, the results of our experiments are the result of generations of the patients introduced in \S~\ref{sec:models}. To stabilize our analysis, we have reported the standard deviations of the results computed over the generations. In the following paragraphs, we exemplify the evaluation process used for each experiment by discussing the error analysis to provide a clear overview of the robustness of the results obtained.

\paragraph{Multiple Choices Question} 
Experiment 1 and Experiment 2, presented in \S~\ref{sec:Typical_Animacy}, are based on a robust pipeline. In the first case, the generation is closely related to a multiple-choice question task. Consequently, the evaluation used heuristics based on string matching between the target choice and the given answer as proposed in \cite{wei2023chainofthought,zheng2024large}. Similar to Experiment 2, where the question is a strict answer \texttt{(Yes) or (No)}. Hence, in this second case, we also used a heuristic based on string matching between the target values and the answer is given. 
Hence, the LLMs were stimulated to generate well-formed defined responses. In Appendix \ref{sec:APP_error1}, we show that the total percentage of responses that did not reflect the defined string-matching heuristics is not sensible and confirms the robustness of the results obtained. In particular, we estimated a maximum misleading response rate of about 0.5\% and 0.6\% (see Table \ref{tab:es_answer_prompt_exp1}) and 2.5-3\% (see Table \ref{tab:es_answer_prompt_exp2}),  which does not affect the final results.  Examples of generation can be seen in Appendix \ref{sec:generation_exp1} and Appendix \ref{sec:generation_exp2}.

\paragraph{Number Generation}
Prompts based on multiple-choice questions or strict answers such as Yes or No are easier to control and analyze. However, in Experiment 3 and Experiment 4, numbers are involved. To manage and control the sensitivity of the prompts, as proposed in Experiments 3 and 4, we added the keyword "\texttt{[num]}" (see \S~\ref{sec:atypical_animacy_exp1} and \S~\ref{sec:atypical_animacy_exp2}). In a similar way, in order to produce a complete and robust analysis, we estimated the final values by profoundly analyzing the numerical outputs or not. We used the Python library \texttt{word2number} to convert the generated literal number into integer values. As displayed in Appendix \ref{sec:APP_error2}, the answers containing literal numbers are significantly minor and do not affect the final evaluations. Finally, the [num] keyword seems to have directed the generation correctly, as reported in the examples shown in Table \ref{tab:es_prompt_exp3} and Table \ref{tab:es_prompt_exp4}.
\end{table}

\begin{table}[]
\section{Models Vesions}
\label{app:model_versions}
\small
\begin{center}
\begin{tabular}{l|c}
\textbf{Model} & \textbf{Version}  \\ 

\hline
\hline
Llama-2-7   &  meta-llama/Llama-2-7b \\
Llama-2-13   &  meta-llama/Llama-2-13b \\
Llama-2-70   &  meta-llama/Llama-2-70b \\ \hline
Mistral-7  & mistralai/Mistral-7B-Instruct-v0.2  \\
Mixtral8x7   & TheBloke/Mixtral-8x7B-Instruct-v0.1-GPTQ  \\  \hline
GPT-3.5-turbo & OpenAI API (gpt-3.5-turbo-0125)  \\
GPT-4 & OpenAI API (gpt-4-1106-preview) \\
\hline
\end{tabular}

\caption{List the versions of the models proposed in this work, which can be found on huggingface.co. We used the configurations described in Appendix \ref{app:model_info} in the repositories for each model *(access to the following models was verified on 1-8-2024).}
\label{tab:versions_models}
\end{center}

\vspace{1cm}

\section{Model and Hyperparameters}
\label{app:model_info}

In our experimental setting, we propose different LLMs: (i) models from the GPT family \cite{openai2023gpt4}: GPT-3.5 (\texttt{gpt-3.5-turbo-0125}) and GPT-4 (\texttt{gpt-4}); (ii) three models from the Llama-2 family \cite{touvron2023llama}: Llama2-7b, Llama2-13b, Llama2-70b, (iii) two models of the MistralAI family: Mistral-7b and Mixtral \cite{jiang2024mixtral}.

In particular, GPTs models are used via API, while for the others, we used versions of the quantized to 4-bit models that use GPTQ (see detailed versions in Table \ref{tab:versions_models})

As discussed in the limitations, our choices are related to reproducibility and the cost associated with non-open-source models.
We use closed-source API and the 4-bit GPTQ quantized version of the model on four 48GB NVIDIA RTXA600 GPUs for all experiments performed only in inference.

Finally, the generation temperature used varies from $\tau = 0$ of GPT models to $\tau = 0.5$ of Llama2s. We choose these temperatures for (mostly) deterministic outputs, with a maximum token length of 256. The other parameters are left unchanged as recommended by the official resources. We will release the code and the dataset upon acceptance of the paper.     

\paragraph{Evaluation}

Finally, as we described in each experiment, we evaluate the accuracy scores.
We compute the string matching between the final answers and the target values. The top-p parameter is set to 1 in all processes and the prompting temperature [0, 1] by repeating the experiments three times.

\end{table}

\begin{table*}[t]
\section{Appendix Error Analysis Strict Answers}
\label{sec:APP_error1}
    \small
    \begin{tabular}{lccccccc}
        \toprule
        \textbf{Type} & GPT-4 & GPT-3.5 & Llama2-70 & Llama2-13 & Llama2-7 & Mixtral & Mistral-7 \\
        \midrule
        Transitive & 0.1\% & 0.1\% & 0.2\% & 0.3\% & 0.4\% & 0.3\% & 0.4\% \\
        Passive & 0.1\% & 0.1\% & 0.2\% & 0.3\% & 0.5\% & 0.2\% & 0.5\% \\
        \bottomrule
    \end{tabular}
        \centering
    \caption{Percentage over 1,000 instances for the \textbf{Transitive} and \textbf{Passive} sub-task (Section \ref{sec:Typical_Animacy_exp}) of generations that do not contain one of the prompted choices. Table \ref{tab:es_answer_prompt_exp1} shows two examples of outputs.}
    \label{tab:error_exp1}

\vskip 10mm

    \begin{tabular}{cccccccc}
        \toprule
        \textbf{Type} & GPT-4 & GPT-3.5 & Llama2-70 & Llama2-13 & Llama2-7 & Mixtral & Mistral-7 \\
        \midrule
        Animated  & 0.5\% & 1\% & 2.5\% & 2.5\% & 3\% & 3\% & 3.5\% \\
        \bottomrule
        Unanimated  & 0.8\% & 1.5\% & 2\% & 2\% & 2.5\% & 3\% & 3\% \\
    \end{tabular}
        \centering
    \caption{Percentage over 4,500 instances for the \textbf{Animated} and \textbf{Unanimated} sub-task of generations that do not contain \texttt{(Yes)} or \texttt{(No)} as explained in Section \ref{sec:Typical_Animacy_exp1.2}. Consequently, it is difficult to assess the answer automatically.}
    \label{tab:error_exp2}

\end{table*}

\begin{table*}[t]
\section{Appendix Error Analysis Numeric Answers}
\label{sec:APP_error2}
    \small
    \begin{tabular}{llccccccc}
        \toprule
        & & GPT-4 & GPT-3.5 & Llama2-70 & Llama2-13 & Llama2-7 & Mixtral & Mistral-7 \\
        \midrule
        \multirow{2}{*}{\textbf{$T_1$}} & Animate & 2(0) & 1(0) & 2(0) & 7(1) & 8(1) & 3(0) & 4(0) \\
         & Inanimate & 2(0) & 4(0) & 5(1) & 7(1) & 8(1) & 4(0) & 5(1) \\
        \midrule
        \multirow{2}{*}{\textbf{$T_2$}} & Animate & 0(0) & 1(0) & 2(0) & 3(0) & 3(0) & 1(0) & 3(0) \\
         & Inanimate & 1(0) & 4(0) & 3(1) & 2(1) & 8(1) & 4(0) & 5(1) \\
        \midrule
        \multirow{2}{*}{\textbf{$T_3$}} & Animate & 0(0) & 0(0) & 1(0) & 1(0) & 1(0) & 1(0) & 1(0) \\
         & Inanimate & 0(0) & 0(0) & 0(0) & 1(0) & 2(0) & 2(0) & 0(0) \\
        \bottomrule
    \end{tabular}
        \centering
    \caption{Number of generations that do not contain numerical values and in brackets that do not contain words meaning numbers. The total instances (sentences) are 60 for each time-step, as introduced in Section \ref{sec:atypical_animacy_exp1}.}
    \label{tab:error_exp3}

\vskip 10mm

    \begin{tabular}{lccccccc}
        \toprule
         & GPT-4 & GPT-3.5 & Llama2-70 & Llama2-13 & Llama2-7 & Mixtral & Mistral-7 \\
        \midrule
         Animate & 2(0) & 3(0) & 7(0) & 8(1) & 8(2) & 4(1) & 5(1) \\
          Inanimate & 1(0) & 3(0) & 6(1) & 8(1) & 9(0) & 2(0) & 5(1) \\
        \midrule
    \end{tabular}
        \centering
    \caption{Number of generations that do not contain numerical values and in brackets that do not contain words meaning numbers. The total number of instances analyzed (sentences) is 60, as introduced in Section \ref{sec:atypical_animacy_exp2}.}
    \label{tab:error_exp4}

\end{table*}

\begin{table*}
    \section{Answers for Experiment 1}
    \label{sec:generation_exp1}
    \noindent
{\setlength{\fboxsep}{-1.4pt} 
\colorbox{lightgray}{ 
\begin{tabular}[t]{|p{0.48\textwidth}|}
    \hline
\texttt{\textbf{Choose which example is acceptable between A and B.}} \\
\texttt{A) Hospitals aren't biked to by the man.
} \\
\texttt{B) Hospitals aren't biked to by the mouth.
} \\
\texttt{\textbf{Answer:}} \\
    \hline
        \texttt{The acceptable example is \textbf{A) Hospitals aren’t biked to by the man.}}\\
\texttt{In English, "biking" typically refers to the activity of riding a bicycle. It is logical to say that a man could ride a bicycle to a hospital. However, it doesn't make sense to say that "the mouth" could bike to a hospital, as the mouth is not capable of such physical activity. Therefore, option A is the appropriate choice because it aligns with the expected action and capability of the subject ("the man").} \\ 
\hline
    \hline
(\textit{answer generated by \textbf{GPT-3.5-turbo}}) \\
\hline
    \end{tabular}
    }
    }
\hfill
{\setlength{\fboxsep}{-1.4pt} 
\colorbox{lightgray}{ 
\begin{tabular}[t]{|p{0.48\textwidth}|}
    \hline
\texttt{\textbf{Choose which example is acceptable between A and B.}} \\
\texttt{A) Hospitals aren't biked to by the man.
} \\
\texttt{B) Hospitals aren't biked to by the mouth.
} \\
\texttt{\textbf{Answer:}} \\
    \hline
        \texttt{
    The correct answer is \textbf{B) Hospitals aren't biked to by the mouth.}} \\
    
\texttt{ Explanation:
In this sentence, "mouth" is a noun that refers to the opening through which food and air pass when we breathe. It doesn't make sense to say that hospitals are biked to by the mouth because hospitals are buildings where medical treatment is provided, and they cannot be reached or accessed through the mouth.}\\
\hline
    \hline
(\textit{answer generated by \textbf{Llama-2-13-chat}}) \\
\hline

    \end{tabular}
    }
    }
\caption{Examples of Answers generated by \textit{GPT-3.5} and \textit{Llama-2-13-chat}. As introduced in Section \ref{sec:Typical_Animacy_exp} and shown in Table \ref{tab:error_exp1}, the answers contain the choices prompted.}
\label{tab:es_answer_prompt_exp1}

\section{Answers for Experiment 2}
\label{sec:generation_exp2}
    \noindent
{\setlength{\fboxsep}{-1.4pt} 
\colorbox{lightgray}{ 
\begin{tabular}[t]{|p{0.48\textwidth}|}
    \hline
\texttt{\textbf{Is the following sentence
plausible? Answer by choosing} (Yes) \textbf{or} (No).} \\
\texttt{\textbf{Sentence:} At the club the cocktails are served by the \uline{tonic}} \\
\texttt{\textbf{Answer:}} \\
    \hline
        \texttt{No, the sentence is not plausible. Cocktails are typically served by bartenders or servers, not by tonic, which is a type of mixer. Tonic water is a beverage ingredient used in cocktails, but it does not serve cocktails itself. Therefore, the sentence does not make sense in the context of cocktail service.}\\
            \hline
\hline
(\textit{answer generated by \textbf{GPT-3.5-turbo}}) \\
\hline
    \end{tabular}
    }
    }
\hfill
{\setlength{\fboxsep}{-1.4pt} 
\colorbox{lightgray}{ 
\begin{tabular}[t]{|p{0.48\textwidth}|}
    \hline
\texttt{\textbf{Is the following sentence
plausible? Answer by choosing (Yes) or (No).}} \\
\texttt{Sentence: At the club the cocktails are served by the \uline{tonic}} \\
\texttt{\textbf{Answer:}} \\
    \hline
        \texttt{Yes, the sentence is plausible. The word "tonic" could refer to a type of alcoholic drink, such as a gin and tonic, which is commonly served at clubs. In this context, the sentence makes sense and is a common way to describe how cocktails are served at a bar or nightclub.}\\
        \hline
    \hline
(\textit{answer generated by \textbf{Llama-2-7-chat}}) \\
\hline

\end{tabular}
    }
    }
\caption{Examples of Answers generated from \textit{GPT-3.5} and \textit{Llama-7-13-chat}. Also in this case, as described in Section \ref{sec:Typical_Animacy_exp1.2}, the prompts are well formed as the output generated with low misleading rates (Table \ref{tab:error_exp2}).}
\label{tab:es_answer_prompt_exp2}

\end{table*}

\begin{table*}

\section{Answers for Experiment 3}
\label{sec:generation_exp3}
    \noindent
    \centering
{\setlength{\fboxsep}{-1.4pt} 
\colorbox{lightgray}{
\begin{tabular}[t]{|p{0.96\textwidth}|}
    \hline
\texttt{Choose a surprising value from 0 to 30 on the following story:
A museum visitor asked the map where the small collection of Van Gogh paintings was located. }\\
\hline
\texttt{Answer: 27 (\textbf{GPT-4})} \\
\texttt{Answer: 22 (\textbf{Mixtral8x7})} \\
\hline
\texttt{\textbf{Given the following story:}}\\
\texttt{A museum visitor asked the map where the small collection of Van Gogh paintings was located. The map was willing to take her there, as there was nothing to do. The museum visitor envied the map because he was allowed to see so much beauty every day. But the map laughingly explained that he always took a Game Boy with him when he was bored. The museum visitor teased the map, joking that he had to be an art barbarian to ever get bored in this museum. }\\
\texttt{\textbf{Choose a surprising value from 0 to 30 on the following story:}
 The map joked that he had seen enough art for the next ten years.} \\
\texttt{\textbf{Answer:[num]}}\\ \hline
    \hline
\texttt{6.} \textit{(answer generated by \texttt{\textbf{GPT-4}})}\\
\texttt{7.
Considering the previous answer, the map's response is less surprising but still slightly unexpected. A surprising value of 7 out of 30 reflects the diminished unexpectedness of the map's attitude, while still acknowledging the earlier context.} \textit{(answer generated by \texttt{\textbf{Mixtral8x7}})}\\
\hline
    \end{tabular}
    }
    }
\hfill

\caption{This example shows the answers provided by \textit{GPT-4} and \textit{Mixtral8x7} using the prompt defined in Section \ref{sec:atypical_animacy_exp1}. The structure of the input and the keyword (\texttt{[num]}) elicit the prompted LLMs to generates a number in digits as output.}
\label{tab:es_prompt_exp3}

\section{Answers for Experiment 4}
\label{sec:generation_exp4}
    \noindent
{\setlength{\fboxsep}{-1.4pt} 
\colorbox{lightgray}{ 
\begin{tabular}[t]{|p{0.96\textwidth}|}
    \hline
\texttt{\textbf{Given the following context:}}\\
\texttt{A girl told a sandwich that an attack was imminent. The sandwich wailed that his family was in danger.
The girl told the sandwich that public places were the most dangerous. The sandwich immediately started calling everyone he knew. }\\
\texttt{\textbf{Choose a surprising value from 0 to 30 on the following story:}
The sandwich was delicious and wanted to make sure
none of his loved ones were in danger.}\\
\texttt{\textbf{Answer:[num]}}\\
\hline
\hline
\texttt{20.} \textit{(answer generated by \texttt{\textbf{GPT-3.5}})}\\
\texttt{18.} \textit{(answer generated by \texttt{\textbf{Mixtral8x7}})}\\
\texttt{15.} \textit{(answer generated by \texttt{\textbf{Llama-2-70-chat}})}\\
\texttt{24.} \textit{(answer generated by \texttt{\textbf{Llama-2-7-chat} and \textbf{Llama-2-13-chat}})}\\
        \hline
\end{tabular}
    }
    }
\caption{This example shows the answers provided by different probed LLMs using the prompt defined in Section \ref{sec:atypical_animacy_exp2}. Also in this experiment the structure of the input and the keyword (\texttt{[num]}) elicit the prompted LLMs to generates a number in digits as output.}
\label{tab:es_prompt_exp4}
    
\end{table*}

\end{document}